\documentclass[10pt,twocolumn,letterpaper]{article}
\newcommand\correspondingauthor{\thanks{Corresponding author. This work is supported by the Sea-NExT Joint Lab}}
\usepackage{cvpr}              

\usepackage{graphicx}
\usepackage{amsmath}
\usepackage{amssymb}
\usepackage{booktabs}
\usepackage{multirow}
\usepackage[dvipsnames]{xcolor}
\usepackage[pagebackref,breaklinks,colorlinks]{hyperref}

\usepackage[capitalize]{cleveref}
\crefname{section}{Sec.}{Secs.}
\Crefname{section}{Section}{Sections}
\Crefname{table}{Table}{Tables}
\crefname{table}{Tab.}{Tabs.}

\newcommand{\wx}[1]{{\color{black}{#1}}}
\newcommand{\wxx}[1]{{\color{black}{#1}}}
\newcommand{\lyc}[1]{{\color{black}{#1}}}
\newcommand{\green}[1]{{\color{OliveGreen}{#1}}}
\newcommand{\pink}[1]{{\color{pink}{#1}}}
\def\cvprPaperID{7878} 
\def\confName{CVPR}
\def\confYear{2022}

\usepackage{overpic}
\usepackage{enumitem} 
\usepackage{overpic} 
\usepackage{color}

\definecolor{turquoise}{cmyk}{0.65,0,0.1,0.3}
\definecolor{purple}{rgb}{0.65,0,0.65}
\definecolor{dark_green}{rgb}{0, 0.5, 0}
\definecolor{orange}{rgb}{0.8, 0.6, 0.2}
\definecolor{red}{rgb}{0.8, 0.2, 0.2}
\definecolor{darkred}{rgb}{0.6, 0.1, 0.05}
\definecolor{blueish}{rgb}{0.0, 0.3, .6}
\definecolor{light_gray}{rgb}{0.7, 0.7, .7}
\definecolor{pink}{rgb}{1, 0, 1}
\definecolor{greyblue}{rgb}{0.25, 0.25, 1}

\usepackage{blindtext}

\renewcommand{\paragraph}[1]{\vspace{1em}\noindent\textbf{#1}.}

\newcommand{\Lapl}{\mathbf{\mathop{\mathcal{L}}}}

\newcommand{\Trans}[1]{{#1}^{\top}}

\newcommand{\Mat}[1]{\textbf{#1}}

\newcommand{\Space}[1]{\mathbb{#1}}
\newcommand{\Set}[1]{\mathcal{#1}}

\begin{document}
\title{Invariant Grounding for Video Question Answering}

\author{Yicong Li$^1$, Xiang Wang$^2$\correspondingauthor, Junbin Xiao$^1$, Wei Ji$^1$, Tat-Seng Chua$^1$\\
$^1$National University of Singapore, 
$^2$University of Science and Technology of China,\\
{\tt\small liyicong@u.nus.edu, xiangwang1223@gmail.com} \\
{\tt\small junbin@comp.nus.edu.sg, jiwei@nus.edu.sg, dcscts@nus.edu.sg}

}

\maketitle

\begin{abstract}
    Video Question Answering (VideoQA) is the task of answering questions about a video.
    At its core is understanding the alignments between visual scenes in video and linguistic semantics in question to yield the answer.
    In leading VideoQA models, the typical learning objective, empirical risk minimization (ERM), \wx{latches on} superficial correlations between video-question pairs and answers \wx{as the alignments}.
    However, ERM can be problematic, \wx{because it tends to over-exploit the spurious correlations between question-irrelevant scenes and answers, instead of inspecting the causal effect of question-critical scenes.}
    As a result, the VideoQA models suffer from unreliable reasoning.
    
    In this work, we first take a causal look at VideoQA and argue that invariant grounding is the key to ruling out the spurious correlations.
    Towards this end, we propose a new learning framework, Invariant Grounding for VideoQA (IGV), to ground the question-critical scene, whose causal relations with answers are invariant across different interventions on the complement.
    With IGV, the VideoQA models are forced to shield the answering process from the negative influence of spurious correlations, which significantly improves the reasoning ability.
    Experiments on three benchmark datasets validate the superiority of IGV in terms of accuracy, visual explainability, and generalization ability over the leading baselines. 
    Our code is available at
    \url{https://github.com/yl3800/IGV}.

    
\vspace{-10pt}
\end{abstract}

\section{Introduction}
\label{sec:intro}

Video Question Answering (VideoQA) \cite{fan2019heterogeneous} is growing in popularity and importance to interactive AI, such as vision-language navigation for in-home robots and personal assistants \cite{DBLP:conf/cvpr/WangHcGSWWZ19,DBLP:conf/cvpr/AndersonWTB0S0G18}.
It is the task of multi-modal reasoning, which answers the natural language question about the content of a given video.
Clearly, inferring a reliable answer requires a deep understanding of visual scenes, linguistic semantics, and more importantly, the visual-linguistic alignments.

Towards this end, a number of VideoQA models have emerged \cite{fan2019heterogeneous,li2019beyond,DBLP:conf/mm/XuZX0Z0Z17,gao2018motionappearance,dang2021hierarchical}.
Scrutinizing these models, we summarize their common paradigm as a combination of two modules:
(1) video-question encoder, which encapsulates the visual scenes of video and the linguistic semantics of question as representations;
and (2) answer decoder, which exploits these representations to model the visual-linguistic alignment and yield an answer.
Consequently, the criterion of empirical risk minimization (ERM) is widely adopted as the learning objective to optimize these modules --- that is, minimizing the loss between the predictive answer and the ground-truth answer.


\begin{figure}[t]
	\centering
	\subcaptionbox{Local mutual information (LMI) between the ``track'' scene and answers.\label{fig:lmi-show}}{
		\includegraphics[width=0.92\linewidth]{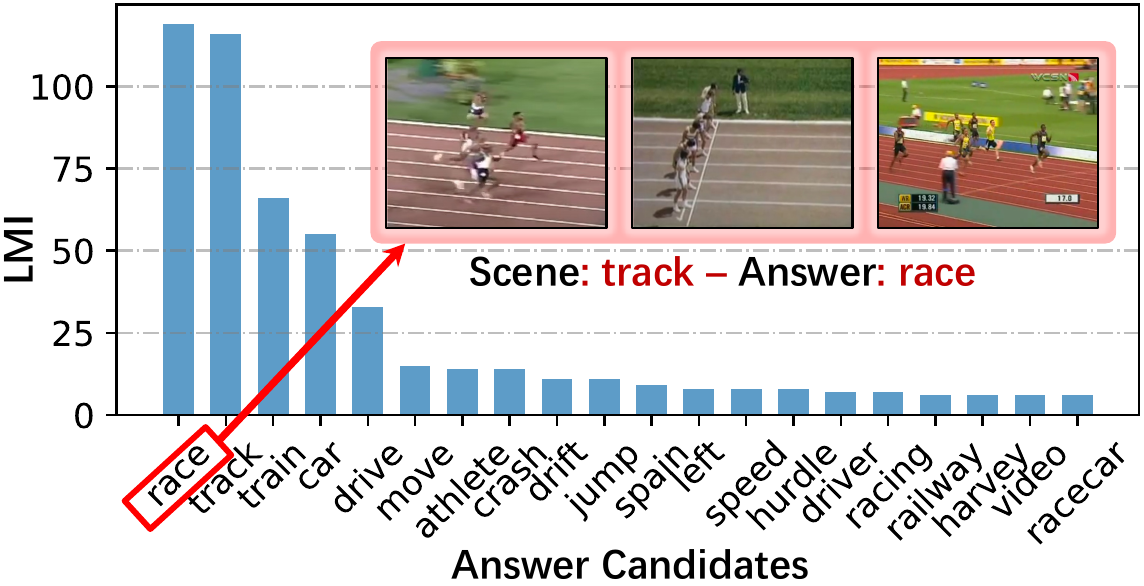}}
	\subcaptionbox{Running example of how the ``track'' complement deviates the answering.\label{fig:runing_example}}{
		\includegraphics[width=0.92\linewidth]{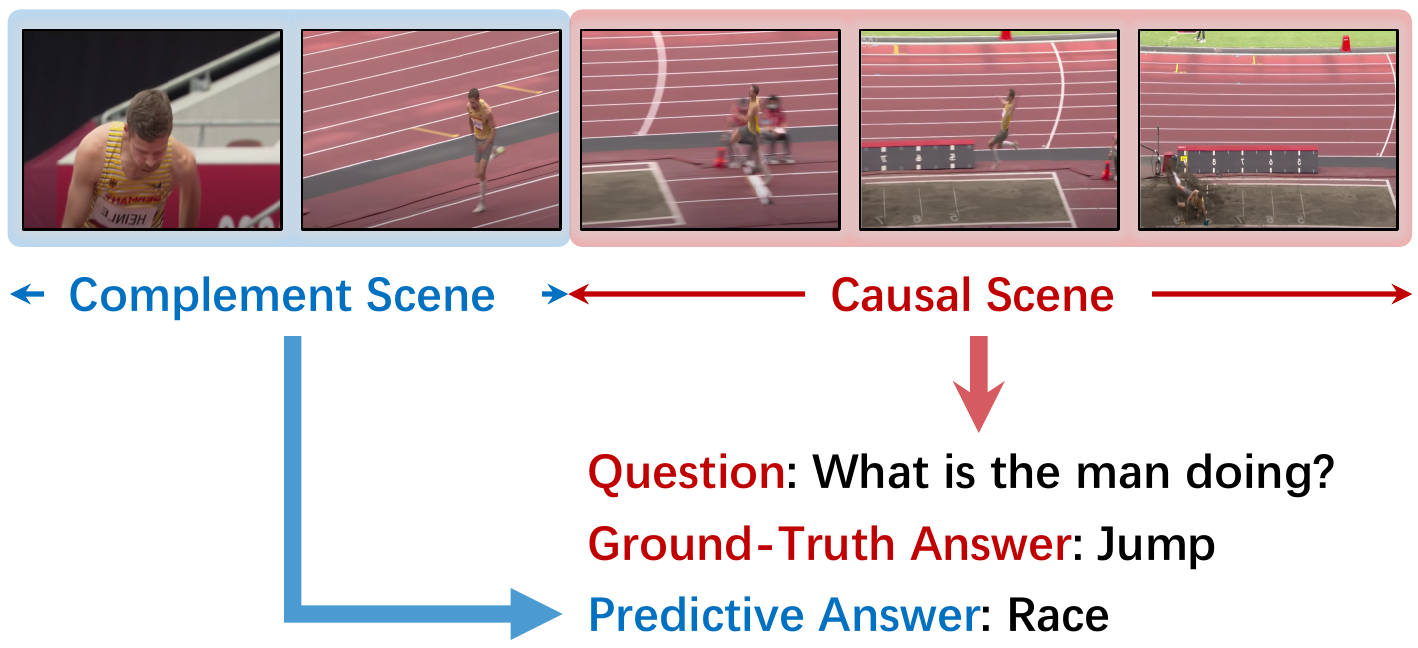}}
	\vspace{-5pt}
	\caption{Running example. (a) Superficial correlations between visual scenes and answers; (b) Suffering from the spurious correlations, VideoQA model fails to answer the question.}
	\label{fig:intro-example}
	\vspace{-15pt}
\end{figure}



However, the ERM criterion is prone to over-exploiting the superficial correlations between video-question pairs and answers.
Specifically, we use the metric of local mutual information (LMI) \cite{schuster2019debiasing} to quantify the correlations between the ``track'' scene and answers.
As Figure \ref{fig:lmi-show} shows, most videos with ``track'' scene are associated with the ``race'' answer.
Instead of inspecting the visual-linguistic alignments (\ie which scene is critical to answer the question), ERM blindly captures all statistical relations.
As Figure \ref{fig:runing_example} shows, it makes VideoQA model naively link the ``track''-relevant videos with the strongly-correlated ``race'' answer, instead of the gold ``jump'' answer.
\wx{Taking a causal look \cite{pearl2000models,pearl2016causal} at VideoQA (see Section \ref{sec:causal-view}), we partition the visual scenes into two parts: (1) causal scene, which holds the question-critical information, and (2) its complement, which is irrelevant to the answer.
We scrutinize that the complement is spuriously correlated with the answer, thus ERM hardly differentiates the effects of causal and complement scenes on the answer.}
Worse still, the unsatisfactory reasoning obstacles the VideoQA model to own the intriguing properties:
\begin{itemize}[leftmargin=*]
\setlength\itemsep{-2pt}
    \item \textbf{Visual-explainability} to exhibit ``Which visual scene are the right reasons for the right answering?'' \cite{DBLP:conf/ijcai/RossHD17,CSS}.
    Taking Figure \ref{fig:runing_example} as an example to answer ``What is the man doing?'', the model should attend the ``jump'' event present in the last three clips, rather than referring to the ``track'' complement in the first two clips.
    One straightforward solution is ``learning to attend'' \cite{xiao2020visual,DBLP:conf/icml/XuBKCCSZB15,DBLP:conf/nips/VaswaniSPUJGKP17} to ground some scenes via the attentive mechanism.
    Nonetheless, guided by ERM, such attentive grounding still suffers from the spurious correlations, thus making the highly-correlated complement grounded.

    \item \textbf{Introspective learning} to double-check ``How would the predictive answer change if the causal scenes were absent?''.
    On top of attentive grounding, the model needs to introspect whether the learned knowledge (\ie attended scene) reliably and faithfully reflects the logic behind the answering. Briefly put, it should fail to answer the question if the causal scenes were removed.
    
    \item \textbf{Generalization ability} to enquire ``How would the predictive answer response to the change of spurious correlations?''.
    As spurious correlations poorly generalize to open-world scenarios, the model should instead latch on the causal visual-linguistic relations that are stable across different environments.
\end{itemize}

Inspired by recent invariant learning \cite{arjovsky2020invariant, wang2021causal,REx}, we conjecture that invariant grounding is the key to distinguishing causal scenes from the complements and overcoming these limitations.
By ``invariant'', we mean that the relations between question-critical scenes and answers are invariant regardless of changes in complements.
Towards this end, we propose a new learning framework, \underline{I}nvariant \underline{G}rounding for \underline{V}ideoQA (\textbf{IGV}).
Concretely, it integrates two additional modules with into the VideoQA backbone model: a grounding indicator, a scene intervener.
Specifically, the grounding indicator learns to attend the causal scenes for a given question and leaves the rest as the complement.
\wxx{Then, we collect visual clips from other training videos to compose a memory bank of complement stratification.}
For the causal part of interest, the scene intervener conducts the causal interventions \cite{pearl2000models,pearl2016causal} on its complement --- that is, replace it with \wxx{the stratification sampled from} the memory bank and compose the ``intervened videos''.
After pairing the casual, complement, and intervened scenes with the question, we feed them into the backbone model to obtain the corresponding predictions:
(1) causal prediction, which approaches the gold answer, so as to achieve visual explainability;
(2) complement prediction, which contains no critical clues to the ground-truth answer, thus enforces the backbone model to perform introspective reasoning;
and (3) intervened prediction, which is consistent with the causal prediction across different intervened complements.
Jointly learning these predictions enables the backbone model to alleviate the negative influence of multi-modal data bias.
It is worthwhile emphasizing that IGV is a model-agnostic strategy, which trains the VideoQA backbones in a plug-and-play fashion.

Our contributions are summarized as follows:
\begin{itemize}[leftmargin=*]
\setlength\itemsep{-2pt}
    \item We highlight the importance of grounding causal scenes from the complements to visual-explainability, generalization, and introspective learning of VideoQA models.
    
    \item We propose a new model-agnostic training scheme, IGV, which incorporates invariant grounding into the VideoQA models, to mitigate the negative influence of multi-modal data bias and enhance the multi-modal reasoning ability.
    
    \item On three benchmark datasets (\ie MSRVTT-QA \cite{DBLP:conf/mm/XuZX0Z0Z17}, MSVD-QA \cite{DBLP:conf/mm/XuZX0Z0Z17}, NExT-QA \cite{DBLP:conf/cvpr/XiaoSYC21}), we conduct extensive experiments to justify the superiority of IGV in training the VideoQA backbones. In particular, IGV significantly outperforms the state-of-the-art models.
\end{itemize}

\section{Preliminaries}
In this section, we summarize the common paradigm of VideoQA models.
Throughout the paper, we denote the random variables and their deterministic values by upper-cased (\eg $V$) and lower-cased (\eg $v$) letters, respectively.

\vspace{5pt}
\noindent\textbf{Modeling}.
Given the video-question pair $(V,Q)$, the primer task of VideoQA is to generate an answer $\hat{A}$ as:
\begin{gather}\label{eq:conventional-modeling}
    \hat{A} = f_{\hat{A}}(V,Q),
\end{gather}
where $f_{\hat{A}}$ is the VideoQA model, which is typically composed of two modules: video-question encoder, and answer decoder.
Specifically, the encoder includes two components:
(1) a video encoder, which encodes visual scenes of the target video as a visual representation, such as motion-appearance memory design \cite{gao2018motionappearance, fan2019heterogeneous}, structural graph representation \cite{jiang2020reasoning, park2021bridge, huang2020locationaware, Wang_2018_ECCV}, hierarchical architecture \cite{le2021hierarchical, dang2021hierarchical};
and (2) a question encoder, which encapsulates linguistic semantics of the question into a linguistic representation, such as global/local representation of textual content \cite{Jiang_Chen_Lin_Zhao_Gao_2020, 2021}, graph representation of grammatical dependencies \cite{park2021bridge}.
On top of these representations, the decoder learns the visual-linguistic alignments to generate the answer.
In particular, the alignments are modeled via cross-modal interaction like graph alignment\cite{park2021bridge}, cross-attention \cite{jiang2020reasoning, li2019beyond,zeng2016leveraging, Jiang_Chen_Lin_Zhao_Gao_2020} and co-memory \cite{gao2018motionappearance}, \etc.

\vspace{5pt}
\noindent\textbf{Learning}.
To optimize these modules, most of the leading VideoQA models \cite{fan2019heterogeneous,gao2018motionappearance,le2021hierarchical,jiang2020reasoning,Jiang_Chen_Lin_Zhao_Gao_2020} cast the multi-modal reasoning problem as a supervised learning task and adopt the learning objective of empirical risk minimization (ERM) as:
\begin{gather}\label{equ:erm-loss}
    \vspace{-30pt}
    \min_{h}\Lapl_{\text{ERM}}(\hat{A}, A),
    \vspace{-30pt}
\end{gather}
where $\Lapl_{\text{ERM}}$ is the risk function to measure the loss between the predictive answer $\hat{A}$ and ground-truth answer $A$, which is usually set as cross-entropy loss \cite{gao2018motionappearance,le2021hierarchical} or hinge loss \cite{fan2019heterogeneous,jiang2020reasoning,DBLP:conf/cvpr/XiaoSYC21}.
In essence, ERM encourages these VideoQA modules to capture the statistical correlations between the video-question pairs and answers.

\section{Causal Look at VideoQA}
\label{sec:causal-view}
From the perspective of causal theory \cite{pearl2000models,pearl2016causal}, we revisit the VideoQA scenario to show superficial correlations between video-question pairs and answers.
We then analyze ERM's suffering from the spurious correlations.

\subsection{Causal Graph of VideoQA}
\begin{figure}[t]
    \centering
    \includegraphics[width=0.9\linewidth]{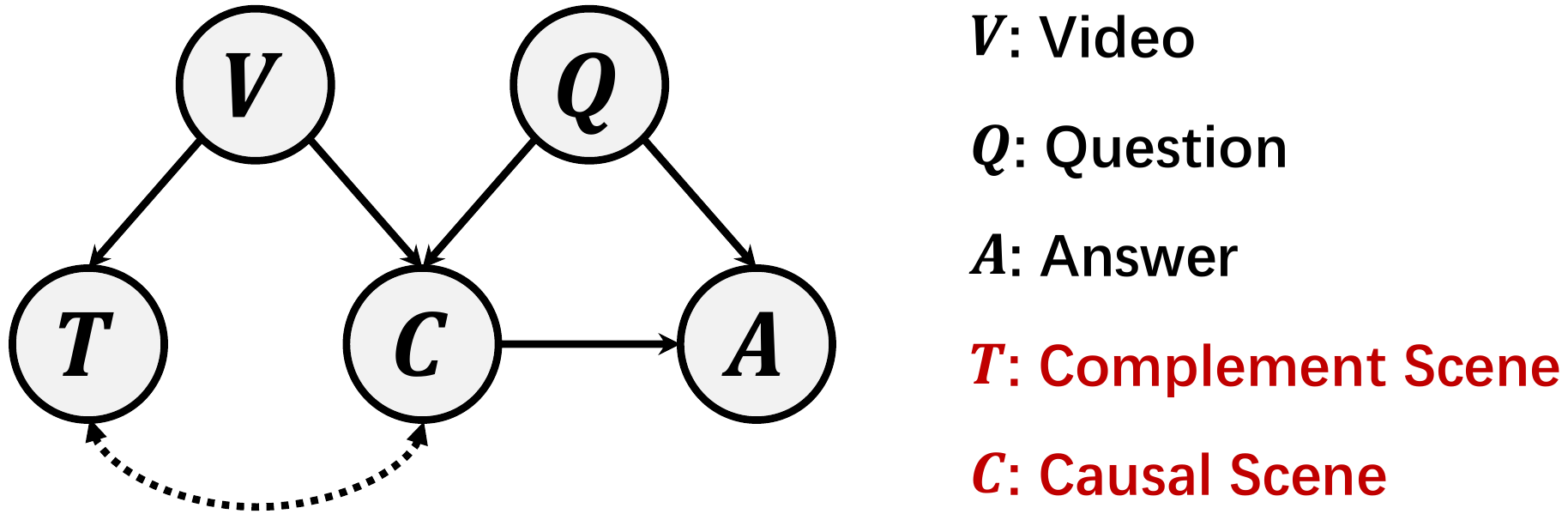}
    \vspace{-5pt}
    \caption{Causal graph of VideoQA}
    \label{fig:scm}
    \vspace{-15pt}
\end{figure}

In general, multiple visual scenes are present in a video.
But only part of the scenes are critical to answering the question of interest, while the rest hardly offers information relevant to the question.
Moreover, the linguistic variations in different questions should activate different scenes of a video.
These facts inspire us to split the video into the causal and complement parts in terms of the question.
Here we use a causal graph \cite{pearl2000models,pearl2016causal} to exhibit the relationships among five variables:
input video $V$, input question $Q$, causal scene $C$, complement scene $T$, ground-truth answer $A$.
Figure \ref{fig:scm} illustrates the causal graph, where each link is a cause-and-effect relationship between two variables:
\begin{itemize}[leftmargin=*]
    \setlength\itemsep{-2pt}
    \item $C\gets V\to T$. The input video $V$ consists of $C$ and $T$. For example, the video in Figure \ref{fig:runing_example} is the combination of the first two clips (\ie $C$) and the last three clips (\ie $T$).
    \item $V\to C\gets Q$. The causal scene $C$ is conditional upon the video-question pair $(V,Q)$, which distills $Q$-relevant information from $V$. For a given $V$, the variations in $Q$ result in different $C$.
    
    \item $Q\to A\gets C$. The answer $A$ is determined by the question $Q$ and causal scene $C$, reflecting the visual-linguistic alignments. Considering the example in Figure \ref{fig:runing_example} again, $C$ is the oracle scene that perfectly explains why ``jump'' is labeled as the ground truth to answer the question.
    \item $T\dashleftarrow\dashrightarrow C$. The dashed arrow summarizes the additional probabilistic dependencies \cite{10.5555/1642718,reason:Pearl09a} between $C$ and $T$.
    Such dependencies are usually caused by the selection bias or inductive bias during the process of data collection or annotation \cite{DBLP:conf/cvpr/TorralbaE11,DBLP:conf/naacl/ChaoHS18}. For example, one mostly collects the videos with the ``jump'' events on the ``track''.
    Here we list three typical scenarios: (1) $C$ is independent of $T$ (\ie $T \bot C$); (2) $C$ is the direct cause of $T$ (\ie $C\rightarrow T$), or vise versa (\ie $C\leftarrow T$); (3) $C$ and $T$ have a common cause $E$ (\ie $C\leftarrow E\rightarrow T$). See Appendix \ref{app:critical-context-example} for details.
\end{itemize}

\vspace{-15pt}
\subsection{Spurious Correlations}
Taking a closer look at the causal graph, we find that the complement scene $T$ and the ground-truth answer $A$ can be spuriously correlated.
Specifically, as the confounder \cite{pearl2000models,pearl2016causal,reason:Pearl09a} between $T$ and $A$, $Q$ and $V$ open the backdoor paths: $T \gets V \to C \to A$ and $T \gets V \to C \gets Q \to A$, which make $T$ and $A$ spuriously correlated even though there is no direct causal path from $T$ to $A$.
Worse still, $T\dashleftarrow\dashrightarrow C$ can amplify this issue.
Assuming $C \to T$, $C$ becomes an additional confounder to yield another backdoor path $T \gets C \to A$.
Such spurious correlations can be summarized as the probabilistic dependence: $A\not\!\bot T$.

As ERM naively captures the statistical correlations between video-question pairs and answers, it fails to distinguish the causal scene $C$ and complement scene $T$, thus failing to mitigate the negative influence of spurious correlations.
As a result, it limits the reasoning ability of VideoQA models, especially in the following aspects:
(1) visual-explainability to reason about ``Which visual scenes are the supporting evidence to answer the question?'';
(2) introspective learning to answer ``How would the answer change if the causal scenes were absent?'';
and (3) generalization ability to enquire ``How would the answer response to the change of spurious correlations?''.

\section{Methodology}

We get inspiration from invariant learning \cite{arjovsky2020invariant, wang2021causal,REx} and argue that invariant grounding of causal scenes is the key to reducing the spurious correlations and overcoming the foregoing limitations.
We then present a new learning framework, \underline{I}nvariant \underline{G}rounding for \underline{V}ideoQA (\textbf{IGV}).

\subsection{Invariant Grounding for VideoQA}

Upon closer inspection on the causal graph, we notice that the ground-truth answer $A$ is independent of the visual complement $T$, only when conditioned on the question $Q$ and the causal scene $C$, more formally:
\begin{gather}\label{equ:invariance}
    A\bot T \mid C,Q.
\end{gather}
This probabilistic independence indicates the invariance --- that is, the relations between the $(C,Q)$ pair and $A$ are invariant regardless of changes in $T$.
The causal relationship $Q\rightarrow A \leftarrow C$ is invariant across different $T$.
Taking Figure \ref{fig:runing_example} as an example, if the question and the causal scene (\ie the last three clips) remain unchanged, the answer should arrive at ``jump'', no matter how the complement varies\footnote{Note that the complement substitutes will not involve the question-relevant scenes, in order to avoid creating additional paths from $T$ to $A$.} (\eg substitute the ``track'' clips by the ``cloud''- or ``sea''-relevant ones).
This highlights that the $(C,Q)$ pair is the key to shielding $A$ from the influence of $T$.

\vspace{5pt}
\noindent\textbf{Modeling.}
However, only the $(V,Q)$ pair and $A$ are available in the training set, while neither $C$ nor the grounding function towards $C$ is known.
This motivates us to incorporate visual grounding into the VideoQA modeling, where the grounded scene $\hat{C}$ aims to estimate the oracle $C$ and guide the prediction of answer $\hat{A}$.
More formally, instead of the conventional modeling (\cf Equation \eqref{eq:conventional-modeling}), we systematize the modeling process as:
\begin{gather}
    \hat{C}=f_{\hat{C}}(V,Q),\quad \hat{A}=f_{\hat{A}}(\hat{C}, Q),
\end{gather}
where $f_{\hat{C}}$ is the grounding model, and $f_{\hat{A}}$ is the VideoQA model that relies on the $(\hat{C},Q)$ pair instead.
See Section \ref{sec:implementations} for our implementations of $f_{\hat{C}}$ and $f_{\hat{A}}$.

\vspace{5pt}
\noindent\textbf{Learning.}
Nonetheless, simply integrating visual grounding with the VideoQA model falls into the ``learning to attend'' paradigm, which still suffers from the spurious correlations and erroneously attends to the complement scenes as $\hat{C}$.
To this end, we exploit the invariance property of $C$ (\cf Equation \eqref{equ:invariance}) and reformulate the learning objective of invariant grounding as:
\begin{gather}
    \min_{f_{\hat{A}},f_{\hat{C}}}\Lapl_{\text{IGV}}(\hat{A}, A),\quad\text{s.t.}~~A\bot \hat{T}\mid\hat{C},Q,
\end{gather}
where $\Lapl_{\text{IGV}}$ is the loss function to our IGV; $\hat{T}=V\setminus\hat{C}$ is the complement of $\hat{C}$.
In the next section, we will elaborate how to implement $\Lapl_{\text{IGV}}$ and achieve invariant grounding.

\subsection{IGV Framework}
\label{sec:implementations}

\begin{figure}[t]
    \centering
    \includegraphics[width=1\linewidth]{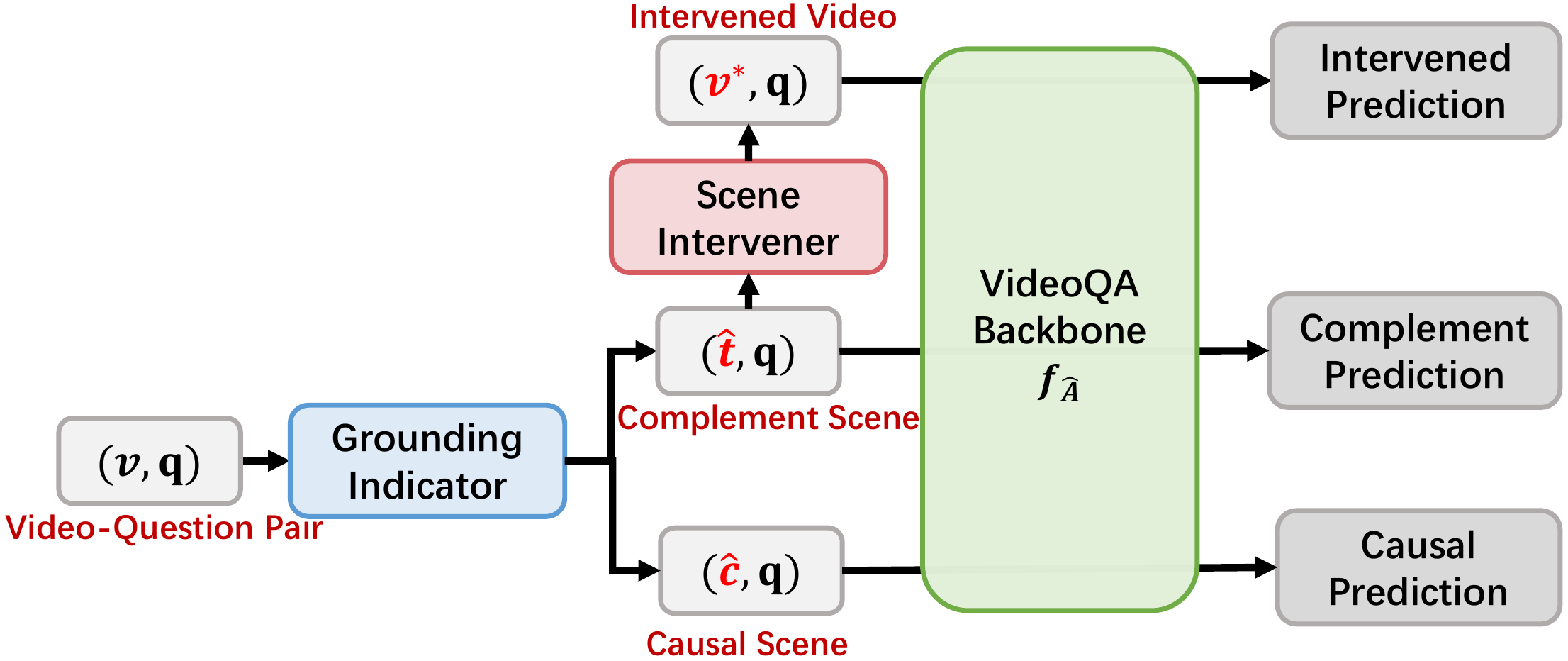}
    \vspace{-15pt}
    \caption{Overview of our IGV framework.}
    \label{fig:overview}
    \vspace{-15pt}
\end{figure}


Figure \ref{fig:overview} displays our IGV framework, which involves two additional modules, the grounding indicator and scene intervener, beyond the VideoQA backbone model $f_{\hat{A}}$.

\subsubsection{Grounding Indicator}
For a video-question pair instance $(v,q)$, at the core of the grounding indicator is to split the video instance $v$ into two parts, $\hat{c}$ and $\hat{t}$, according to the question $q$.
Towards this end, it first employs two independent LSTMs \cite{10.1162/neco.1997.9.8.1735} to encode the visual and linguistic characteristics of $v$ and $q$, respectively:
\begin{gather}\label{equ:vl-lstm}
    \Mat{v}_{g},\Mat{v}_{l} = \text{LSTM}_{1}(v),\quad \Mat{q}_{g},\Mat{q}_{l} = \text{LSTM}_{2}(q),
\end{gather}
where the features of $v$ are $K$ fixed visual clips, while $q$ is associated with $L$ language tokens;
LSTM$_1$ outputs $\Mat{v}_{l}\in\Space{R}^{K\times d}$ as the local representations of clips, and yields the last hidden state $\Mat{v}_{g}\in\Space{R}^{d}$ as the global representations of the holistic video.
Analogously, LSTM$_2$ generates $\Mat{q}_{l}\in\Space{R}^{L\times d}$ as the local representations of tokens, and makes the last hidden state $\Mat{q}_{g}\in\Space{R}^{d}$ represent the question holistically, here $d$ is the hidden dimension.

Upon these representations, the attention scores are constructed to indicate the importance of each visual clip.
Here we devise $\Mat{p}_{\hat{c}}\in\Space{R}^{K}$ to exhibit the probability of each clip belonging to the causal scene $\hat{c}$, while $\Mat{p}_{\hat{t}}\in\Space{R}^{K}$ is in contrast to $\Mat{p}_{\hat{c}}$ to show how likely each clip composes the complement $\hat{t}$.
The formulations are as follows:
\begin{align}
    \Mat{p}_{\hat{c}} &= \text{Softmax}(\text{MLP}_{1}(\Mat{v}_{l})\cdot\Trans{\text{MLP}_{2}(\Mat{q}_{g})}),\\
    \Mat{p}_{\hat{t}} &= \text{Softmax}(\text{MLP}_{3}(\Mat{v}_{l})\cdot\Trans{\text{MLP}_{4}(\Mat{q}_{g})}),
\end{align}
where four multilayer perceptrons (MLPs) are employed to distill useful information:
$\text{MLP}_{1}(\Mat{v}_{l}),\,\text{MLP}_{3}(\Mat{v}_{l})\in\Space{R}^{K\times d'}$,  $\text{MLP}_{2}(\Mat{q}_{g}),\,\text{MLP}_{4}(\Mat{q}_{g})\in\Space{R}^{d'}$
; $d'$ is the feature dimension.
However, as the soft masks make $\hat{c}$ and $\hat{t}$ overlapped, the attentive mechanism cannot shield the answering from the influence of the complement.
Hence, the grounding indicator produces discrete selections instead to make $\hat{c}$ and $\hat{t}$ disjoint.
Nonetheless, simple sampling or selection is not differentiable.
To achieve differentiable discrete selection, we apply Gumbel-Softmax \cite{DBLP:conf/iclr/JangGP17}:
\begin{gather}
    \Mat{I} = \text{Gumbel-Softmax}([\Mat{p}_{\hat{c}},\Mat{p}_{\hat{t}}]),
\end{gather}
where Gumbel-Softmax is built upon the concatenation of $\Mat{p}_{\hat{c}}$ and $\Mat{p}_{\hat{t}}$ (\ie $[\Mat{p}_{\hat{c}},\Mat{p}_{\hat{t}}]\in\Space{R}^{K\times 2}$), and outputs the indicator vector $\Mat{I}\in\Space{R}^{K\times2}$ whose first and second column indexes $\hat{c}$ and $\hat{t}$ over k clips, respectively.
As such, we can devise $\hat{c}$ and $\hat{t}$ as follows:
\begin{gather}
    \hat{c} = \{I_{k0}\cdot v_{k}|I_{k0}=1\},\quad \hat{t} = \{I_{k1}\cdot v_{k}|I_{k1}=1\},
\end{gather}
where $I_{0k}$ and $I_{1k}$ suggests that the $k$-th clip belongs to the causal and complement scenes, respectively.


\subsubsection{Scene Intervener}
\begin{figure}[t]
    \centering
    \includegraphics[width=0.96\linewidth]{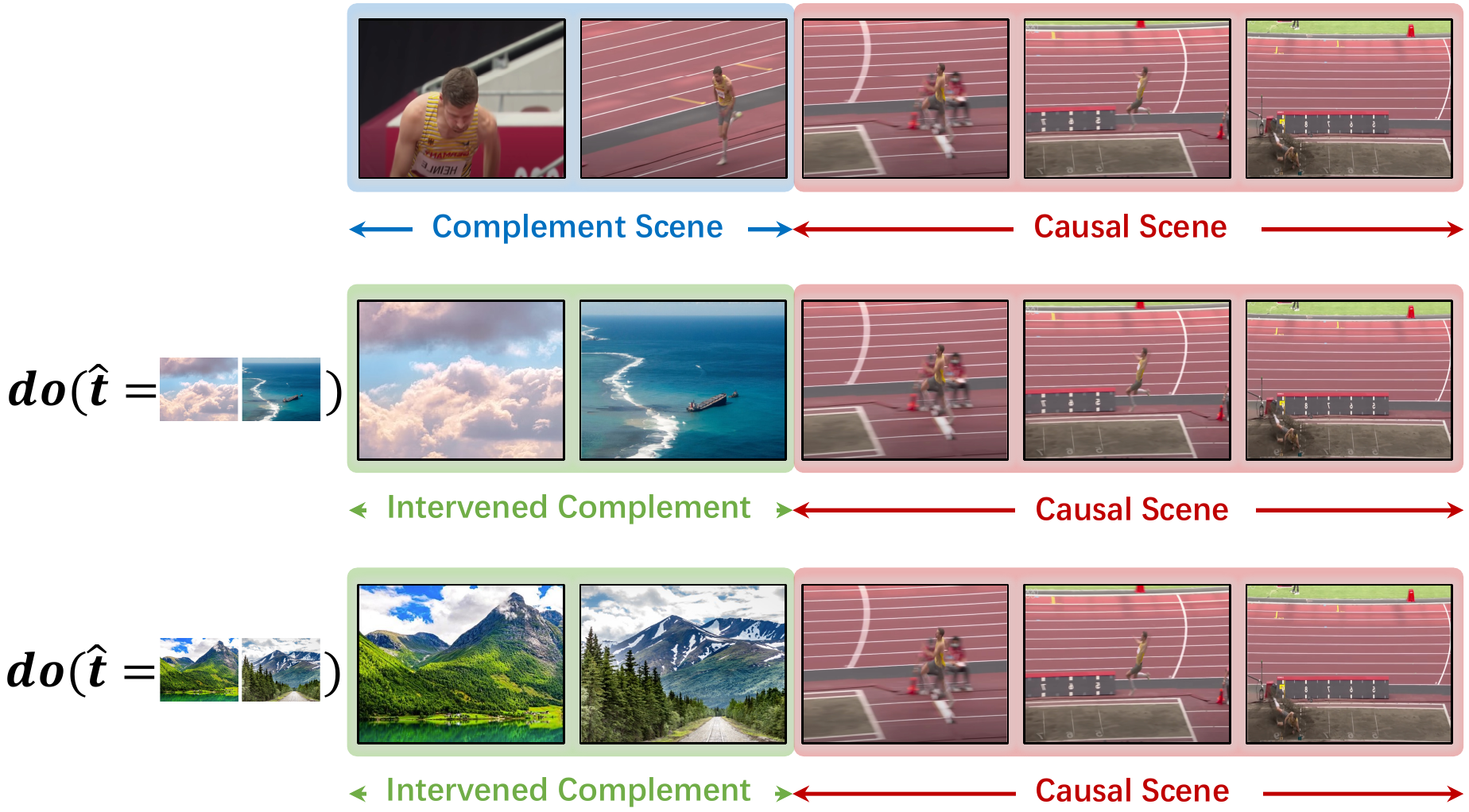}
    \vspace{-5pt}
    \caption{Illustration of interventional distribution.}
    \label{fig:intervene_exp}
    \vspace{-15pt}
\end{figure}


It is challenging to learn the grounding indicator, \wxx{owing to the lack of supervisory signals of clip-level importance.}
To remedy this issue, we propose the scene intervener, which preserves the estimated causal scene $\hat{c}$ but intervenes the estimated complement $\hat{t}$ to create the ``intervened videos'', as Figure \ref{fig:intervene_exp} shows.

Specifically, for the observed video-question pairs during training, \wxx{the scene intervener first collects visual clips from other training videos as a memory bank of complement stratification,} $\hat{\Set{T}}=\{\hat{t}\}$.
Then, for the video of interest $v=\hat{c}\cup\hat{t}$, the intervener conducts causal interventions \cite{pearl2000models,pearl2016causal} on its $\hat{t}$ --- that is, random sample a complement stratification $\hat{t}^{*}\in\hat{\Set{T}}$ to replace $\hat{t}$ and combine it with $\hat{c}$ at hand as a new video $v^{*}=\hat{c}\cup\hat{t}^{*}$.

It is worthwhile mentioning that, distinct from the current invariant learning studies \cite{arjovsky2020invariant, wang2021causal,REx} that only partition the training set into different environments, our scene intervener exploits the interventional distributions \cite{DBLP:conf/aaai/TianKP06} instead.
The interventional distribution (\ie, the videos with the same interventions) can be viewed as one environment.

\subsubsection{VideoQA Backbone Model}
Inspired by \cite{jiang2020reasoning}, we design a simple yet effective architecture
as our backbone predictor,
\lyc{where the video encoder is shared with the grounding indicator.}
It embodies convolutional graph networks (GCN) to propagate clip-level visual messages, then integrates cross-modal fused local and global representations via BLOCK fusion \cite{BenYounes_2019_AAAI}.
See Appendix \ref{app:backbone} for the detailed architecture. 

\subsubsection{Joint Training}
For a video-question pair instance $(v,q)$, we have established the causal scene $\hat{c}$, complement scene $\hat{t}$, and intervened video $v^{*}$ via the grounding indicator and scene intervener.
Pairing them with $q$ synthesizes three new instances: $(\hat{c},q)$, $(\hat{t},q)$, $(v^{*},q)$.
We next feed these instances into the backbone VideoQA model $f_{\hat{A}}$ to obtain three predictions:
\begin{itemize}[leftmargin=*]
    \setlength\itemsep{-2pt}
    \item \textbf{Causal prediction}. As the causal scene $\hat{c}$ is expected to be sufficient and necessary to answer the question $q$, we leverage its predictive answer $f_{\hat{A}}(\hat{c},q)$ to approach the ground-truth answer $a$ solely:
    \begin{gather}
        \Lapl_{\hat{c}} = \text{XE}(f_{\hat{A}}(\hat{c},q), a),
    \end{gather}
    where XE denotes the cross-entropy loss.
    
    \item \textbf{Complement prediction}. As no critical clues should exist in the complement scene $\hat{t}$ to answer the question $q$, we encourage its predictive answer $f_{\hat{A}}(\hat{t},q)$ to evenly predict all answers.
    This uniform loss is formulated as:
    \begin{gather}
        \Lapl_{\hat{t}} = \text{KL}(f_{\hat{A}}(\hat{t},q), u),
    \end{gather}
    where KL denotes KL-divergence, and $u$ is the uniform distribution over all answer candidates.
    
    \item \textbf{Intervened prediction}. According to the invariant constraint (\cf Equation \eqref{equ:invariance}), the causal relationship between the causal scene and the answer is stable across different complements.
    To parameterize this constraint, we enforce all $v$'s intervened versions to hold the consistent predictions:
    \begin{gather}
        \Lapl_{v^{*}} = \Space{E}_{\hat{t}^{*}\in\Set{T}}(\text{KL}(f_{\hat{A}}(v^{*},q), f_{\hat{A}}(\hat{c},q))).
    \end{gather}
    \vspace{-25pt}
\end{itemize}
Aggregating the foregoing risks, we attain the learning objective of IGV:
\begin{gather}
    \Lapl_{\text{IGV}} = \Space{E}_{(v,q,a)\in\Set{O}^{+}}\Lapl_{\hat{c}} + \lambda_{1}\Lapl_{\hat{t}} + \lambda_{2}\Lapl_{v^{*}},
\end{gather}
where $\Set{O}^{+}$ is the training set of the video-question pair $(v,q)$ and the ground-truth answer $a$; $\lambda_{1}$ and $\lambda_{2}$ are the hyper-parameters to control the strengths of invariant learning.
Jointly learning these predictions enables the VideoQA backbone model to uncover the question-critical scene, so as to mitigate the negative influence of spurious correlations between the question-irrelevant complement scene and answer.
In the inference phase, we use the causal prediction $f_{\hat{A}}(\hat{c},q)$ to answer the question.

\section{Experiments}
We conduct extensive experiments to answer the following research questions:
\begin{itemize}[leftmargin=*]
\setlength\itemsep{-.20em}
    \item \textbf{RQ1:} How effect is IGV in training VideoQA backbones as compared with the State-of-the-Art (SoTA) models?
    \item \textbf{RQ2:} How do the loss component and feature setting affect the performance?
    \item \textbf{RQ3:} What are the learning patterns and insights of IGV training?
\end{itemize}
\noindent\textbf{Settings:}
We compare IGV with seven baselines from families of Memory, GNN and Hierarchy (Appendix \ref{app:baseline}) on three VideoQA datasets: 
\textbf{NExT-QA} \cite{DBLP:conf/cvpr/XiaoSYC21} which features causal and temporal action interactions among multiple objects. It contains about 47.7K manually annotated questions for multi-choice QA collected from 5.4K videos with an average length of 44s.
\textbf{MSVD-QA} \cite{DBLP:conf/mm/XuZX0Z0Z17} and \textbf{MSRVTT-QA} \cite{DBLP:conf/mm/XuZX0Z0Z17} are two prevailing datasets that focus on the description of video elements. They respectively contain 50K and 243K QA pairs with open answer space over 1.6K and 6K. For all three datasets, we follow their official data splits for experiments and report accuracy as evaluation metric. 

\vspace{5pt}
\noindent\textbf{Implementation Details:}
For the visual feature, we follow previous works \cite{le2021hierarchical, jiang2020reasoning, DBLP:conf/cvpr/XiaoSYC21} and extract video feature as a combination of motion and appearance representations by using the pre-trained 3D ResNeXt-101 and ResNet-101, respectively. Specifically, each video is uniformly sampled into $K$=16 clips, where each clip is represented by a combined feature vector $v_{k}^{d_v}$, where $d_v$ equals 4096. Similar to \cite{DBLP:conf/cvpr/XiaoSYC21}, we obtain the contextualized word representation from the finetuned BERT model, and the feature dim $d_q$ is 768.  For our model, the dimension of the hidden states are set to $d$ = 512, and the number of graph layers in
IGV backbone predictor is 2. During training, IGV is optimized by Adam optimizer with the initial learning rate of 1e-4, which will be halved if no validation improvements in 5 epochs. We set the batch size to 256 and a maximum of 60 epochs. (See Appendix \ref{app:implementation} for more details and complexity analysis). 

\vspace{5pt}

\subsection{Main Results (RQ1)}\label{section:rq1}
\subsubsection{Comparisons with SoTA Methods}
\setlength{\tabcolsep}{9pt}
\begin{table}[t]
\small
  \centering
  \caption{Comparison of accuracy on NExT-QA test set.  The \textbf{best} and \underline{second-best} results are highlighted.}
    \vspace{-5pt}
    \begin{tabular}{l|cccc}
    \toprule
    Models & Causal     & Temp     & Descrip     & All \\
    \midrule
    \midrule
    Co-Mem \cite{gao2018motionappearance} & 45.85 & \underline{50.02} & 54.38 & 48.54 \\
    HCRN \cite{le2021hierarchical} & 47.07 & 49.27 & 54.02 & 48.82 \\
    HME  \cite{fan2019heterogeneous} & 46.76 & 48.89 & 57.37 & 49.16 \\
    HGA \cite{jiang2020reasoning}  & \underline{48.13} & 49.08 & \underline{57.79} & \underline{50.01} \\
    \midrule
    IGV(Ours)  & \textbf{48.56} & \textbf{51.67} & \textbf{59.64} & \textbf{51.34} \\
    Abs. Improve & +0.43 & +1.65 & +1.85 & +1.33 \\
    \bottomrule
    \end{tabular}
  \label{tab:sota-next}%
  \vspace{-15pt}
\end{table}%

\setlength{\tabcolsep}{5pt}
\begin{table}[t]
\small
  \centering
  \caption{Comparison of accuracy on MSVD-QA and MSRVTT-QA test set. "$\dagger$" indicates the result is re-implementation with the publicly available code}
     \vspace{-5pt}
    \begin{tabular}{ll|cc}
    \toprule
   \multicolumn{2}{c|}{Models} & MSVD-QA  & MSRVTT-QA \\
    \midrule
    \midrule
    \multirow{3}{*}{Memory}
   &  AMU \cite{DBLP:conf/mm/XuZX0Z0Z17}   & 32.0    & 32.0  \cr
   &    HME \cite{fan2019heterogeneous}  & 33.7  & 33.0 \cr
   &    Co-Mem$\dagger$ \cite{gao2018motionappearance} & 34.6  & 35.3 \cr
    \midrule
    \multirow{2}{*}{GNN}
   & HGA$\dagger$ \cite{jiang2020reasoning}  & 35.4  & 36.1 \cr
   & B2A  \cite{park2021bridge} & 37.2  & \underline{36.9} \cr
    \midrule
    \multirow{2}{*}{Hierarchy} 
   & HCRN \cite{le2021hierarchical}  & 36.1  & 35.6 \cr
  &  HOSTR \cite{dang2021hierarchical} & \underline{39.4}  & 35.9 \cr
    \midrule
    \multirow{2}{*}{Causal view} 
    & IGV (Ours)  & \textbf{40.8} & \textbf{38.3} \cr
    & Abs. Improve & +1.4 & +1.4 \cr
    \bottomrule
    \end{tabular}
  \label{tab:sota-ms}%
  \vspace{-15pt}
\end{table}%

As shown in Table \ref{tab:sota-next} and Table \ref{tab:sota-ms}, our method outperforms SoTAs with questions of all sub-types surpassing their competitors. Specifically, we have two major observations: 

    First, on NExT-QA, IGV gains remarkable improvement on \textit{temporal} type (+1.65\%), the underlying explanation are: 1) \textit{temporal} question generally corresponds to video content with a longer time span, which requires more introspective grounding of the causal scene. Fortunately, IGV's design philosophy comfort such demand by wiping out the trivial scenes, which takes up a huge proportion in \textit{temporal} type, thus making the predicting faithful.  2) \textit{temporal} questions tend to include a temporal indicative phase (\eg "\textit{at the end of the video}") that serves as a strong signal for grounding indicator to locate the target window.
    
    Second, along with \textit{descriptive} questions on NExT-QA, the result on MSRVTT-QA and MSVD-QA (both emphases on question of \textit{descriptive} type) demonstrate the superiority in \textit{descriptive} question across all three datasets (+1.85\% on NExT-QA, +1.4\% on MSRVTT-QA and MSVD-QA). Such improvement is underpinned by logic that answering \textit{descriptive} questions requires scrutiny on the scene of interest, instead of a holistic view of the entire sequence. Accordingly, targeted prediction inducted by IGV concentrates reasoning on keyframes, thus achieving better performance. 
    As a consequence, such improvement strongly validates that IGV generalizes better over various environments.
    \vspace{-10pt}

 
\subsubsection{Backbone Agnostic}\label{section:model-agnostic}
By nature, our IGV principle is orthogonal to backbone design, thus helping to boost any off-the-shelf SoTAs without compromising the underlying architecture.
We therefore experimentally testify the generality and effectiveness of our learning strategy by marrying the IVG principle with methods from two different categories: Co-Mem \cite{gao2018motionappearance} from memory-based architecture and HGA \cite{jiang2020reasoning} from Graph-based method. 
Table \ref{tab:model-agnostic} shows the results on three backbone predictors (including ours). Our findings are:

    \textbf{1. Better improvement for severe bias.}
    We notice that the improvement on MSVD-QA (+3.1\%$\sim$4.7\%) is considerably larger than that on MSRVTT-QA (+1.4\%$\sim$2\%). Such expected discrepancy is caused by the fact that, although identical in question type, MSRVTT-QA is almost 5 times larger than MSVD-QA (\#QA pairs 243K \textit{vs} 50K). As a result, the baseline model trained on MSRVTT-QA is gifted with better generalization ability, whereas the model on MSVD-QA still suffers from severe shortcut correlation. For the same reason, the IGV framework achieves much better improvement in the severe-shortcut situation (\eg MSVD-QA). Such discrepancy validates our motivation of eliminating statistic dependency.
    
    \textbf{2. Constant improvement for each method.} 
    Through row-wise inspection, we notice that for each benchmark, IGV can bring considerable improvement across different backbone models (+3.1\%$\sim$4.7\% for MSVD-QA, +1.4\%$\sim$2\% for MSRVTT-QA). Such stable enhancement strongly verifies our modal-agnostic statement.
\setlength{\tabcolsep}{7pt}
\begin{table}[t]
\small
  \centering
  \caption{IGV strategy is applied to different SoTAs methods. "+IGV" denoted our strategy is incorporated.}
   \vspace{-5pt}
    \begin{tabular}{l|cc|cc}
    \toprule
    \multirow{2}[2]{*}{Models} & \multicolumn{2}{c}{MSVD-QA} & \multicolumn{2}{c}{MSRVTT-QA} \\
          & Baseline & +IGV & Baseline & +IGV \\
    \midrule
    \midrule
    Co-Mem \cite{gao2018motionappearance} & 34.6  & \textbf{37.7} & 35.3  & \textbf{37.3} \\
    HGA \cite{jiang2020reasoning}  & 35.4  & \textbf{38.8} & 36.1  & \textbf{37.5} \\
    Our Backbone  & 36.1  & \textbf{40.8} & 36.3  & \textbf{38.3} \\
    \bottomrule
    \end{tabular}
  \label{tab:model-agnostic}%
  \vspace{-10pt}
\end{table}%


\begin{figure*}[t]
  \centering
  \includegraphics[width=0.95\textwidth]{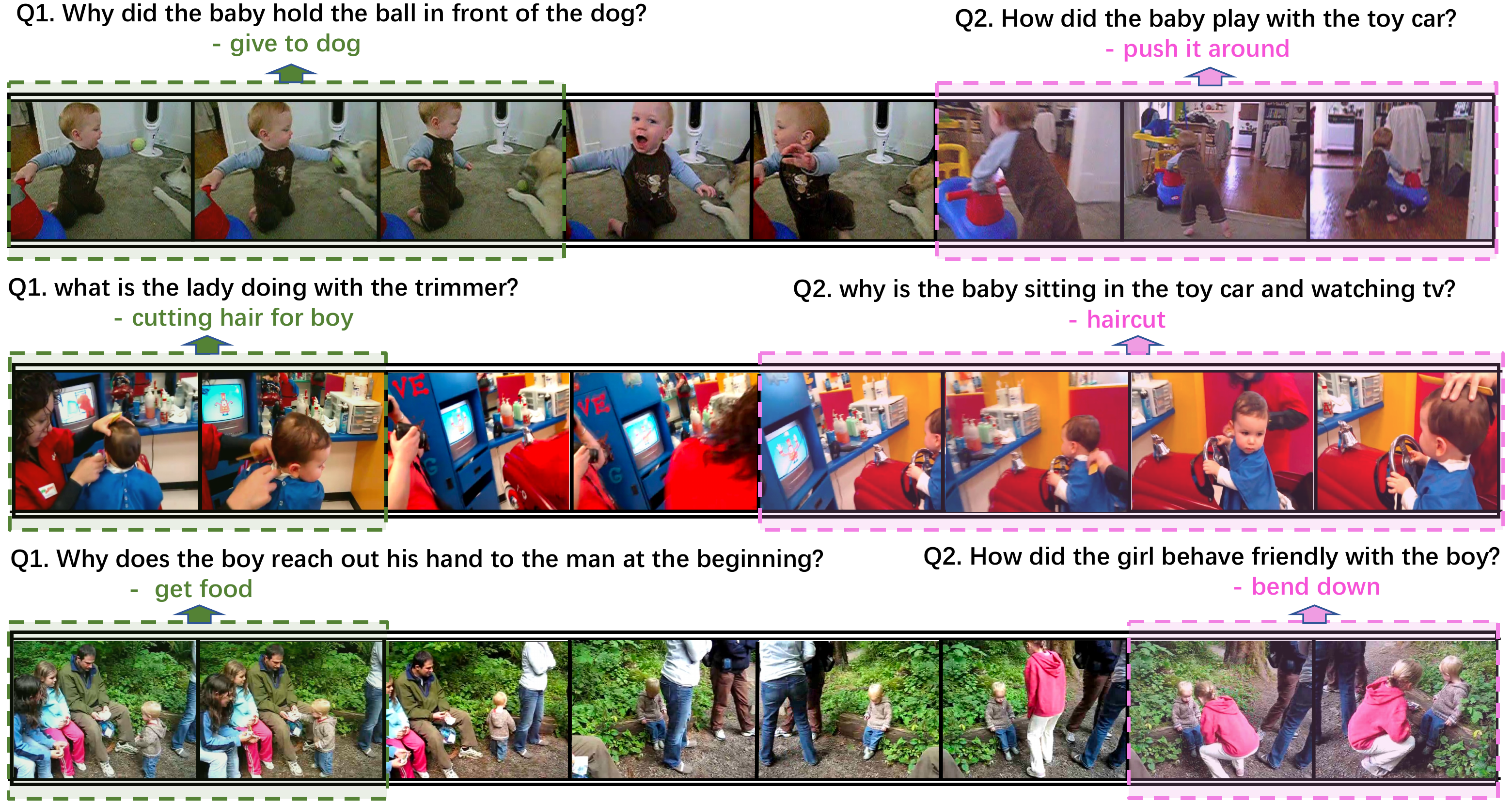}
  	\vspace{-0.07in}
  \caption{Visualization of grounding result on the correct prediction cases from NExT-QA. Each video comes with two questions that demand causal scene of different time span.  The \green{green} and \pink{pink} windows indicate the causal scenes for the corresponding questions.}
  \label{fig:case-study}
 	\vspace{-13pt}
\end{figure*}

\subsection{In-Depth Study (RQ2)}\label{section:rq2}
\subsubsection{Contributions of Different Loss Components}

An in-depth comprehension of IGV framework requires careful scrutiny on its components. Alone this line, we exhaust the combination of IGV loss components and design three variants: $\Lapl_{\hat{c}},$ $\Lapl_{\hat{c}}+\Lapl_{\hat{t}}$ and $\Lapl_{\hat{c}}+\Lapl_{v^{*}}$.
Table \ref{tab:ablation-loss} shows the result of the above variants on two benchmarks across two backbone predictors. Our observations are as follow:
\begin{itemize}[leftmargin=*]
\setlength\itemsep{-.20em}
    \item \wxx{Using $\Lapl_{\hat{c}}$ solely, which can be viewed as a special case of ERM-guided attention}, hardly outperforms the baseline, because grounding indicators can not identify the causal scene without clip-level supervision. Such an expected result reflects our motivation in interventional design. 
    \item $\Lapl_{\hat{c}}+\Lapl_{\hat{t}}$ and $\Lapl_{\hat{c}}+\Lapl_{v^{*}}$ matched equally in accuracy that consistently surpass baseline and $\Lapl_{\hat{c}}$ in all cases. Such progress shows the effectiveness of intervention strategy and introspective regularization imposed on complement.
    \item In all cases, $\Lapl_{\hat{c}}+\Lapl_{\hat{t}}+\Lapl_{v^{*}}$ further boosts the performance significantly, which shows $\Lapl_{\hat{t}}$ and $\Lapl_{v^{*}}$ contribute in different aspects and their benefits are mutually reinforcing. 
    \vspace{-25pt}
\end{itemize}
\setlength{\tabcolsep}{6.5pt}
\begin{table}[tbp]
\small
  \centering
  \caption{Study of IGV loss components}
   \vspace{-5pt}
    \resizebox{0.99\linewidth}{!}{
    \begin{tabular}{l|cc|cc}
    \toprule
    \multirow{2}[1]{*}{Variants} & \multicolumn{2}{c}{MSVD-QA} & \multicolumn{2}{c}{MSRVTT-QA} \\
          & Our Backbone   & Co-Mem \cite{gao2018motionappearance} & Our Backbone   & Co-Mem \cite{gao2018motionappearance} \\
    \midrule
    \midrule
    Baseline & 36.1  & 34.6  & 36.3  & 35.3 \\
    $\Lapl_{\hat{c}}$     & 36.0    & 33.3  & 36.7  & 36.0 \\
    $\Lapl_{\hat{c}}+\Lapl_{\hat{t}}$   & 37.4  & 36.1  & 37.8  & 36.8 \\
    $\Lapl_{\hat{c}}+\Lapl_{v^{*}}$   & 38.2  & 36.3  & 37.4  & 36.2 \\
    $\Lapl_{\hat{c}}+\Lapl_{\hat{t}}+\Lapl_{v^{*}}$ & \textbf{40.8} & \textbf{37.7} & \textbf{38.3} & \textbf{37.3} \\
    \bottomrule
    \end{tabular}
    }
  \label{tab:ablation-loss}%
  \vspace{-10pt}
\end{table}%

\subsubsection{Study of Feature}
\vspace{-4pt}
By convention, we study the effect of the input condition by ablation on the visual feature. Particularly, we denote \textbf{APP} for tests that adopt only appearance feature as input and  \textbf{MOT} for tests that utilize motion feature alone. Figure \ref{fig:feat_analysis} delivers results on two benchmarks, where we observe:

First, IGV can improve the performance significantly for all input conditions, which generalizes the effectiveness of our framework. Similar to Table \ref{tab:model-agnostic}, the improvement on MSVD-QA is larger than that on MSRVTT-QA, which solidifies our finding in Section \ref{section:model-agnostic}.

Second, compared to motion feature, IGV brings distinctively larger improvements using appearance feature. Considering the causal nature of IGV, we conclude that static correlation tends to bias more in appearance feature. 
\vspace{-5pt}

\subsubsection{Study of Hyper-parameter}
\vspace{-3pt}

\begin{figure}[t]
    \centering
    	\subcaptionbox{\label{fig:feat_analysis}}{
		\includegraphics[width=0.48\linewidth]{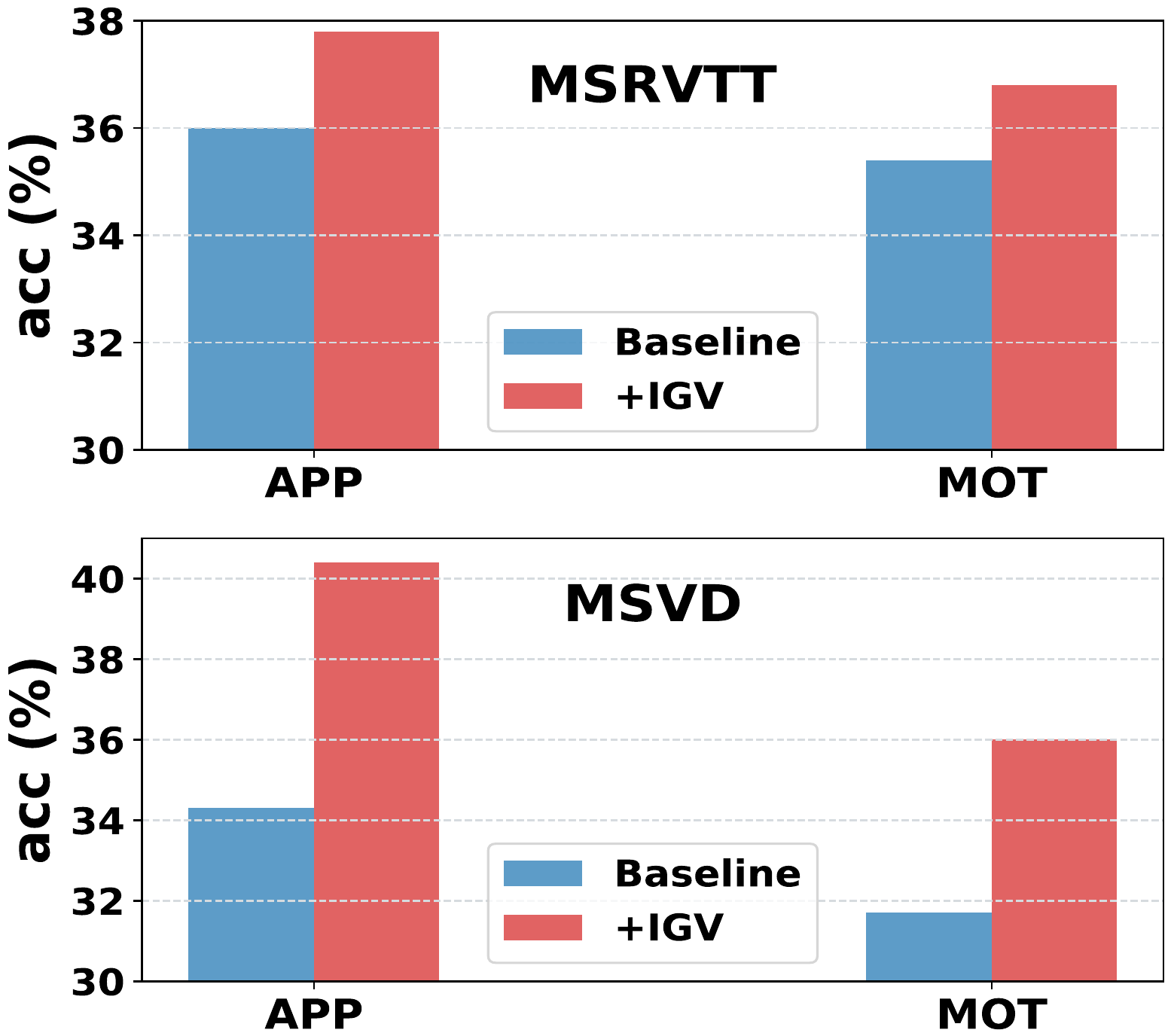}}
		\subcaptionbox{\label{fig:lambda}}{
	    \includegraphics[width=0.48\linewidth]{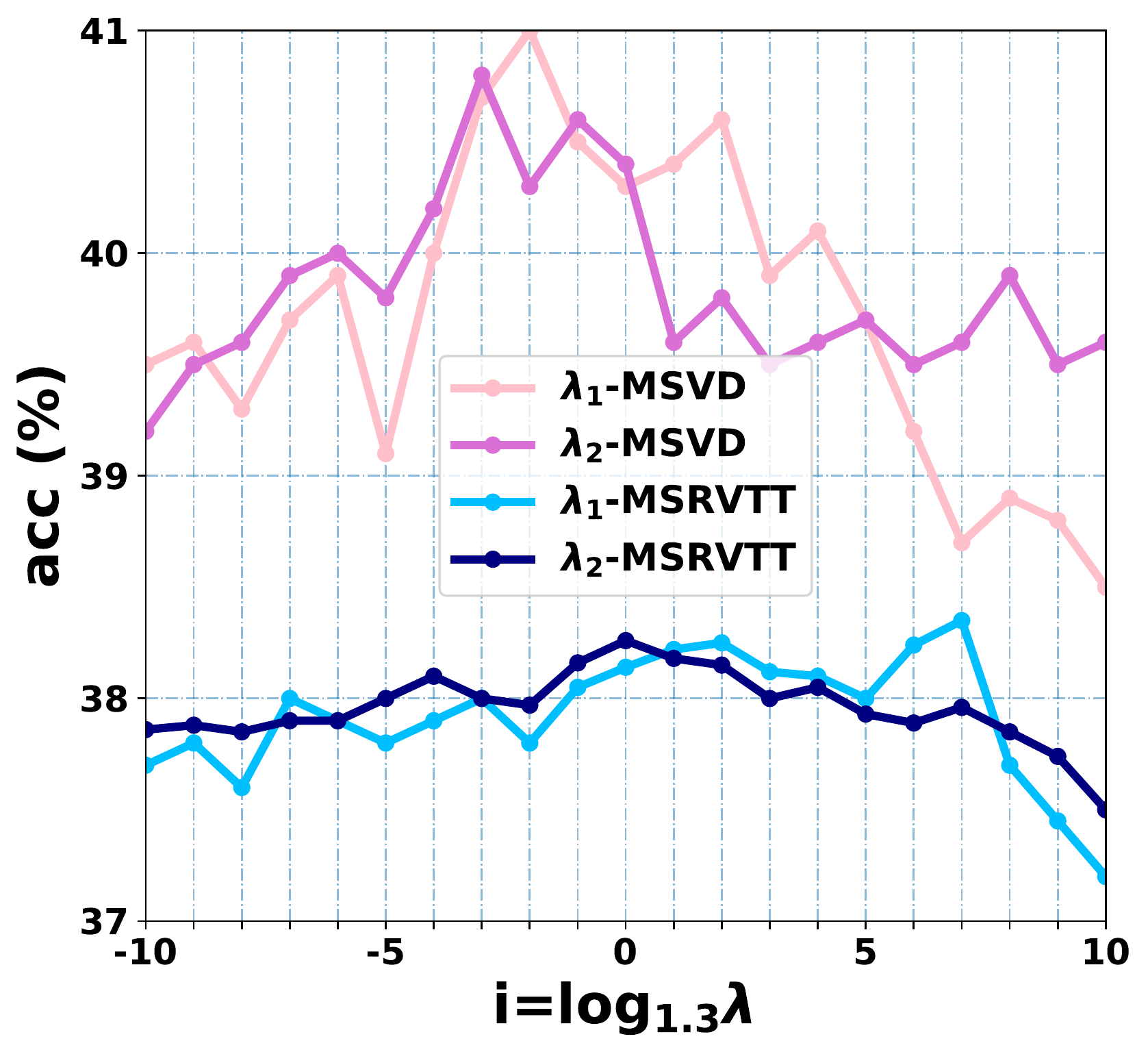}}
    \vspace{-8pt}
    \caption{(a) Study of feature setting.; (b) Study of $\lambda_1$ and $\lambda_2$.}
    \vspace{-15pt}
\end{figure}

To validate the sensitivity of IGV against the hyper-parameters, we conduct experiments with variations of $\lambda_1$ and $\lambda_2$ on two datasets. Without loss of generalization, we tune $\lambda_1$ ($\lambda_2$) as sample of $\left\{ 1.3^{i} \mid -10\le i\le 10, \; i \in\mathbb{Z}  \right\}$,  while keeping the $\lambda_2$ ($\lambda_1$) as 1. According to Figure \ref{fig:lambda}, we have follow observations:

For MSVD-QA, we observe consistent peaks around 0.8 for both $\lambda_1$ and $\lambda_2$. Comparatively, fluctuation on MSRVTT-QA is more moderate, where tuning on $\lambda$ only causes a 1.5\% difference in their accuracy. It's noteworthy that IGV outperforms the baseline by a large margin (+3\%) under all tests, which indicates IGV's robustness against variation of hyper-parameters. Additionally, comparing to $\lambda_2$, IGV is more sensitive to $\lambda_1$. Typically, the performance suffers a drastic degradation for $\lambda_1$ larger than 5 on both datasets. Whereas $\lambda_2$ maintain above 39\% (MSVD-QA) and 37.5\% (MSRVTT-QA) for all tests.

\subsection{Qualitative analysis (RQ3)}\label{section:rq3}
As mentioned in Section \ref{sec:intro} , IGV is empowered with visual-explainability, and is apt to account for the right scene for its prediction.  Following this essence, we grasp the learning insight of IGV by inspecting some correct examples from the NExT-QA dataset and show the visualization in Figure \ref{fig:case-study}. Concretely, each video comes with two questions that emphasize different parts of the video. We notice that, even for the same video, our grounding window is question-sensitive to enclose the explainable content with correct prediction. Nonetheless, we also observe results of \textit{insufficient-grounding} on the 
third row Q2, where the girl starts to bend down before the last two frames, even though the most informative last two frames are encompassed. 


\section{Related works}
\label{sec:related}
\noindent\textbf{Video Question Answering (VideoQA).}
Aiming to answer the question in a video scenario, VideoQA is defined as an escalation of imageQA, because the temporal nature of the input has enriched its reasoning process as well as the answer space. Previous efforts towards VideoQA establish their contribution on either a better multi-modal interaction or stronger video representation. Specifically, early studies tend to impose sophisticated cross-modal fusion via attention \cite{zeng2016leveraging,li2019beyond,jiang2020reasoning} or dynamic memory \cite{gao2018motionappearance, DBLP:conf/mm/XuZX0Z0Z17, fan2019heterogeneous}, while more recent approaches perform relation reasoning through visual or textual graph \cite{jiang2020reasoning,huang2020locationaware,park2021bridge}. In addition, current efforts that model video as a hierarchical structure also intrigue wide interest. Among them, HCRN \cite{le2021hierarchical} stack conditional relation blocks in different feature granularity, whereas HOSTR \cite{dang2021hierarchical} employs a spatio-temporal graph for multilevel reasoning. Despite their effectiveness, their visual-explainability still dwells on ERM-guided attention weights, which only reflect the intensity of feature-prediction correlation.

\vspace{5pt}
\noindent\textbf{Invariant Learning.} 
Multi-modal datasets tend to display inherent bias in some forms \cite{DBLP:journals/corr/abs-1811-05013, DBLP:conf/naacl/ThomasonGB19, DBLP:conf/emnlp/RohrbachHBDS18}. In contrast to overarching reality, the collection process \cite{DBLP:conf/cvpr/TorralbaE11, DBLP:conf/naacl/ChaoHS18} degrades its generalization ability by introducing undesirable correlations between the inputs and the ground truth annotations. 

To overcome such correlation, invariant learning is developed to discover causal relations from the causal factors to the response variable, which remains constant across distributions. 
As the most prevailing formulation, IRM \cite{arjovsky2020invariant} promotes this philosophy from feature level to representation level by finding a data representation $\Upsilon$, from which the optimal predictor $\varphi$ can yield the prediction $\Upsilon\circ \varphi$ that is stable across all environments.
In terms of environment acquisition, previous studies either manually partition the training set by prior knowledge \cite{DBLP:conf/cvpr/AndersonWTB0S0G18}, or generates data partition iteratively via adversarial environment inference \cite{DBLP:conf/icml/CreagerJZ21,wang2021causal}. Our method, instead of partitioning the training, assumes no prophets about environments but performs causal intervention to perturb the original distribution. To the best of our knowledge, IGV is the first work that introduces invariant learning as a model-agnostic framework to 
VideoQA.
\vspace{-3pt}

\section{Conclusions}
\vspace{-5pt}
In this paper, we pinpoint that the spurious visual-linguistic correlations in VideoQA are triggered by question-irrelevant scenes.
We propose a novel invariant grounding framework, IGV, to distinguish the causal scene and emphasize its causal effect on the answer.
With the grounding indicator and scene intervener, IGV captures the causal patterns that remain stable across complements.
Extensive experiments verify the effectiveness of IGV on different backbone VideoQA models.

Our future work includes two aspects: 1) the spurious correlations can nest in entities, object-level invariant learning is promising to alleviate this issue; 
2) as the current intervention strategy might threaten the causal prediction by introducing complement with new shortcuts, we will explore new intervention methods.



\clearpage

{
    \small
    \bibliographystyle{ieee_fullname}
    \bibliography{macros,main}
}

\appendix

\setcounter{page}{1}

\twocolumn[
\centering
\Large
\textbf{Invariant Grounding for Video Question Answering} \\
\vspace{0.5em}Supplementary Material \\
\vspace{1.0em}
] 
\appendix
\section{Example of context type}
\label{app:critical-context-example}
\begin{figure}[t]
    \centering
    \includegraphics[width=0.96\linewidth]{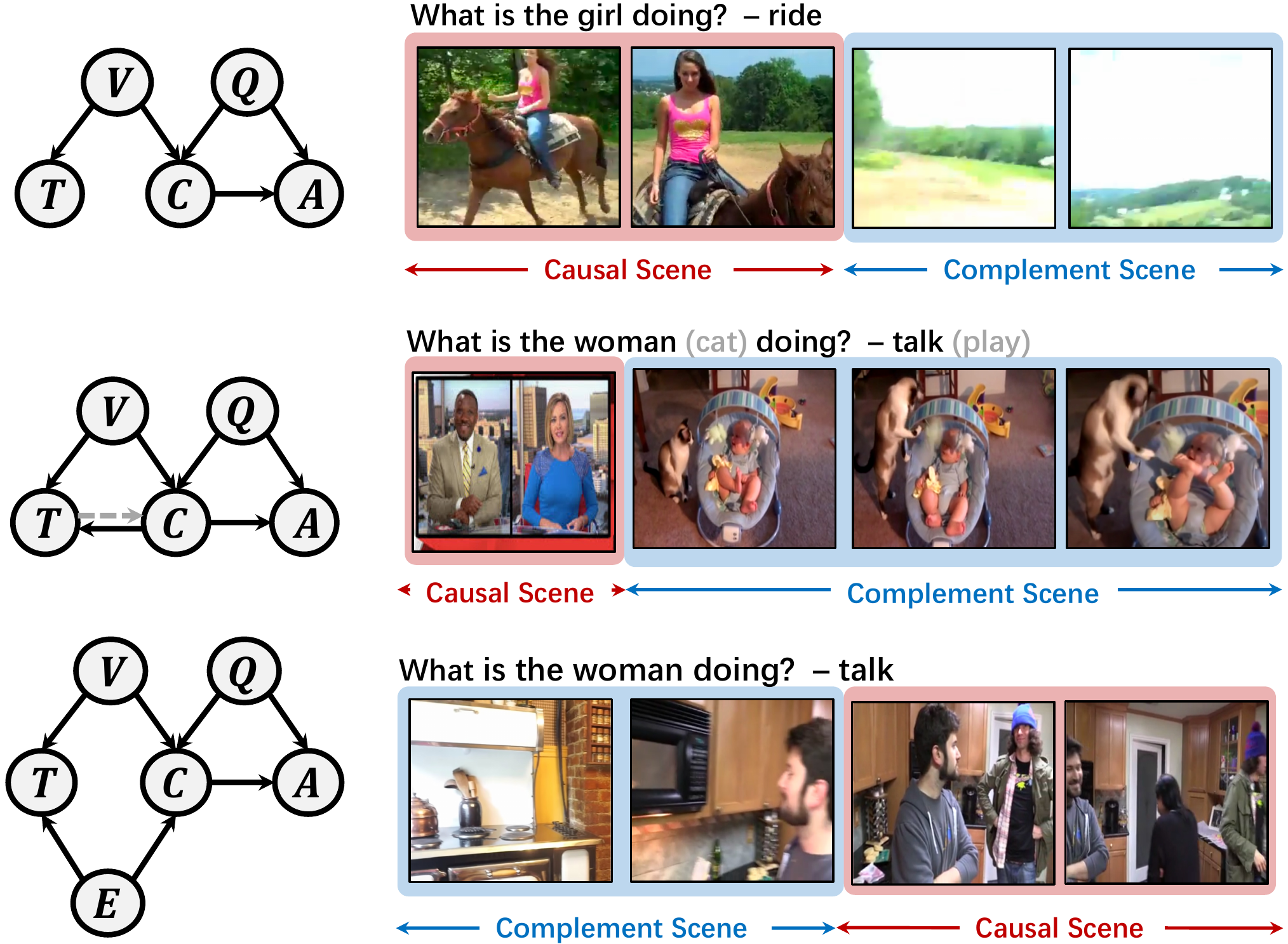}
    \caption{Illustration of complement type.}
    \label{fig:complement type}
\end{figure}

As shown in Figure \ref{fig:complement type}, we classify the relation between causal scene and its complement (\eg $T\dashleftarrow\dashrightarrow C$) into three types, where each row encompasses a causal graph (left) that depicts typical causal-complement relation demonstrated in the example (right):
\begin{itemize}[leftmargin=*]
\setlength\itemsep{-.20em}
    \item In the first row, $C$ and $T$ has no causal relation (\ie $T \bot C$). 
    \item The second row shows a scenario that $C$ is the direct cause of $T$ (\ie $C\rightarrow T$), or vise versa if the question is modified (\eg 'What is the cat doing?')  
    \item Similar to the example in Figure \ref{fig:intro-example}, the third row demonstrates how shortcut deviate the prediction from the gold answer (\eg "talk") to false prediction (\eg "cook") via common cause $E$ (\eg visual concept "kitchen") since LMI between visual concept "kitchen" and candidate answer "cook" is much higher than it is with "talk".  
\end{itemize}

\section{Our backbone}
\label{app:backbone}
Most VideoQA architectures from the state of the art are compatible with our IGV learning strategy. To testify, we design a simple and effective architecture inspired by \cite{jiang2020reasoning}. Specifically, $f_{\hat{A}}$ is presented as a combination of a visual-question mixer and an answer classifier. The mixer first encode $\hat{c}$:
\begin{gather}
    \Mat{v}_{g}^{\hat{c}}, \Mat{v}_{l}^{\hat{c}}=\text{LSTM}_{5}(\hat{c})
\end{gather}
where outputs $\Mat{v}_{g}^{\hat{c}} \in \mathbb{R}^{d}$, $\Mat{v}_{l}^{\hat{c}} \in \mathbb{R}^{N \times d}$ denote the global and local feature of $\hat{c}$ respectively. Then, based on the concatenation of local representation $\Mat{q}_{l}$ (\cf Equation \eqref{equ:vl-lstm}) and $\Mat{v}_{l}^{\hat{c}}$ , we construct an undirected heterogeneous graph that propagates information over each video shot and each question token. Typically, the adjacency matrix $\mathcal{G}_{\hat{c}} \in \mathbb{R}^{(L+N)\times (L+N)}$ is computed as the node-wise correlation scores in form of dot-product similarity, where $N\le K$ is the sequence length of casual scene. The output of the graph is assembled as holistic local factor $\Mat{s}^{\hat{c}}_l \in \mathbb{R}^{d}$ via a attention pooling operator.
More Formally, the process is as follows: 
\begin{gather}
    \Mat{x}_{\hat{c}}=[\Mat{v}_{l}^{\hat{c}};\Mat{q}_{l}], \;\; \mathcal{G}_{\hat{c}}=\sigma(\text{MLP}_{5}(\Mat{x}_{\hat{c}}))\cdot\Trans{\sigma(\text{MLP}_{6}(\Mat{x}_{\hat{c}}))}
\end{gather}
\begin{gather}
    \Mat{z}_{\hat{c}}=\text{GCN}(\Mat{x}_{\hat{c}}, \mathcal{G}_{\hat{c}})
\end{gather}
\begin{gather}
    \Mat{s}_{\hat{c}}^l=\text{Pooling}(\Mat{z}_{\hat{c}})
\end{gather}
where $\Mat{x}_{\hat{c}}$, $\Mat{z}_{\hat{c}}= \in \mathbb{R}^{(L+N)\times d}$ 
denote the input and output of graph reasoning, $\text{MLP}_5$ and $\text{MLP}_6$ denote is affine projection followed by ReLU activation $\sigma(\cdot )$. 
To capture the global information, our mixer integrates two global factors $\Mat{v}_{g}^{\hat{c}}$ and $\Mat{q}_{g}$ into holistic representation via BLOCK fusion \cite{BenYounes_2019_AAAI}:
\begin{gather}
    \Mat{s}_{\hat{c}}^g=\text{Block}(\Mat{v}_{g}^{\hat{c}},\Mat{q}_{g})
\end{gather}
Similarly, we obtain the final representation by applying the BLOCK again to global and local factor, which is further decoded into answer space with classifier $\Psi$:
\begin{gather}
    \Mat{s}_{\hat{c}}=\text{Block}(\Mat{s}_{\hat{c}}^g,\Mat{s}_{\hat{c}}^l)
\end{gather}
\begin{equation}
    \hat{y}_{\hat{c}}=\Psi(\Mat{s}_{\hat{c}})
\end{equation}
Analogously, we can obtain the predictive answer for $\hat{t}$ and $v^{*}$ via the shared backbone predictor. 

\section{Baselines}
\label{app:baseline}
We compare our design against some existing work, which can be categorized into three categories: 
1) \textbf{Memory-based} methods that perform multi-step reasoning via updating the recurrent unit,  which refines the cross-modal representation iteratively. Specifically, AMU \cite{DBLP:conf/mm/XuZX0Z0Z17}, Co-Mem\cite{gao2018motionappearance} apply this module to encode the  visual representation, and HME \cite{fan2019heterogeneous} managed better exploitation for both modalities; 
2) \textbf{Graph-based} methods like HGA \cite{jiang2020reasoning} and B2A \cite{ park2021bridge} adopt graph reasoning on the clip-level, whose adjacent matrix is built on node-wise visual similarity. Comparatively, B2A additionally establishes a text graph through question parsing, and abridge two modalities via message passing;
3) \textbf{Hierarchical-based} methods HOSTR \cite{dang2021hierarchical} and HCRN \cite{le2021hierarchical} have similar hierarchical conditional architectures. Their discrepancy lies in the feature granularity, where HCRN grounds the temporal relation between frames, while HOSTR roots in object trajectories.

\section{Implementation details}
\label{app:implementation}
All experiments are conducted on GPU NVIDIA Tesla V100 installed on Ubuntu 18.0.4. 
In terms of complexity, our algorithm matched equally with the corresponding baseline.  As a comparison, the default backbone model is trained for 2 hours till convergence on MSRVTT-QA, whereas IGV
takes 2.6 hours. For space complexity, since we use the same predictor for the causal, complement, and intervened prediction, IGV only takes 10\% more parameters than the default backbone model. 



\end{document}